\documentclass[sigconf]{acmart}
\usepackage{balance}
\usepackage{subfigure}
\usepackage{threeparttable}
\usepackage{colortbl}

\newcommand{\ra}[1]{\renewcommand{\arraystretch}{#1}}
\newcommand{\midsepremove}{\aboverulesep = 0mm \belowrulesep = 0mm}
\newcommand{\midsepdefault}{\aboverulesep = 0.605mm \belowrulesep = 0.984mm}
\newcommand{\crule}[3][black]{\textcolor{#1}{\raisebox{-.25\height}{\rule{#2}{#3}}}}

\definecolor{colorborder}{HTML}{F3F3F3}
\definecolor{twhite}{HTML}{FFFFFF}
\definecolor{colornotmarked}{HTML}{FCFCFC}
\definecolor{colormarked}{HTML}{2E75B6}
\definecolor{colormarkedlight}{HTML}{dae8f6}
\definecolor{colorgreyedout}{HTML}{777777}
\definecolor{colorscientific}{HTML}{D2DEEF}
\definecolor{colorindustry}{HTML}{9DC3E6}
\definecolor{colorconcept}{HTML}{EAEFF7}
\definecolor{colorverylow}{HTML}{F3FFEF}
\definecolor{colorlow}{HTML}{E5F5E0}
\definecolor{colormedium}{HTML}{A1D99B}
\definecolor{colorhigh}{HTML}{31A354}

\newcommand{\tm}{\cellcolor{colormarked} \color{white}}
\newcommand{\tml}{\cellcolor{colormarkedlight} \color{white}}

\newcommand{\tsa}{\cellcolor{colorscientific}}
\newcommand{\tia}{\cellcolor{colorindustry}}
\newcommand{\tc}{\cellcolor{colorconcept}}
\newcommand{\tca}{\cellcolor{colorlow}}
\newcommand{\tcb}{\cellcolor{colormedium}}
\newcommand{\tcc}{\cellcolor{colorhigh}}
\newcommand{\tcz}{\cellcolor{colorverylow}}

\definecolor{tableblue}{HTML}{709BBF}
\definecolor{theader}{HTML}{6DAFFF}
\definecolor{tablegrid}{HTML}{B6CBDC}
\definecolor{tlightgrey}{HTML}{EEEEEE}
\definecolor{tblack}{HTML}{000000}
\definecolor{tdarkorange}{HTML}{F56B00}
\definecolor{torangelight}{HTML}{FFCE93}
\definecolor{tmandarin}{HTML}{FFC702}
\definecolor{tmandarinlight}{HTML}{FFD7C2}

\definecolor{tred}{RGB}{255,59,48}
\definecolor{tredl}{RGB}{255,177,172}
\definecolor{torange}{RGB}{255,149,0}
\definecolor{torangel}{RGB}{255,213,153}
\definecolor{tyellow}{RGB}{255,204,0}
\definecolor{tyellowl}{RGB}{255,235,153}
\definecolor{tgreen}{RGB}{76,217,100}
\definecolor{tgreenl}{RGB}{183,240,193}
\definecolor{ttealblue}{RGB}{90,200,250}
\definecolor{ttealbluel}{RGB}{189,233,253}
\definecolor{tblue}{RGB}{0,122,255}
\definecolor{tbluel}{RGB}{153,202,255}
\definecolor{tpurple}{RGB}{8,86,214}
\definecolor{tpurplel}{RGB}{156,187,239}
\definecolor{tpink}{RGB}{255,45,85}
\definecolor{tpinkl}{RGB}{255,171,187}

\def\BibTeX{{\rm B\kern-.05em{\sc i\kern-.025em b}\kern-.08emT\kern-.1667em\lower.7ex\hbox{E}\kern-.125emX}}
    
\setcopyright{acmlicensed}

\graphicspath{{figures/}{pictures/}{images/}{./}} % where to search for the images

\begin{document}

\fancyhead{}
  % do not delete this code.

% The "title" command has an optional parameter, allowing the author to define a "short title" to be used in page headers.
\title{Video-based Analysis of Soccer Matches}

% The "author" command and its associated commands are used to define the authors and their affiliations.
% Of note is the shared affiliation of the first two authors, and the "authornote" and "authornotemark" commands
% used to denote shared contribution to the research.

\author{Maximilian T. Fischer}
\affiliation{%
	\institution{University of Konstanz}
	\streetaddress{Universitätsstrasse 10}
	\city{Konstanz}
	\country{Germany}}
\email{Max.Fischer@uni-konstanz.de}

\author{Daniel A. Keim}
\affiliation{%
	\institution{University of Konstanz}
	\streetaddress{Universitätsstrasse 10}
	\city{Konstanz}
	\country{Germany}}
\email{Daniel.Keim@uni-konstanz.de}

\author{Manuel Stein}
\affiliation{%
	\institution{University of Konstanz}
	\streetaddress{Universitätsstrasse 10}
	\city{Konstanz}
	\country{Germany}
}
\email{Manuel.Stein@uni-konstanz.de}

%
% By default, the full list of authors will be used in the page headers. Often, this list is too long, and will overlap
% other information printed in the page headers. This command allows the author to define a more concise list
% of authors' names for this purpose.

%
% The abstract is a short summary of the work to be presented in the article.
\begin{abstract}
	With the increasingly detailed investigation of game play and tactics in invasive team sports such as soccer, it becomes ever more important to present causes, actions and findings in a meaningful manner. Visualizations, especially when augmenting relevant information directly inside a video recording of a match, can significantly improve and simplify soccer match preparation and tactic planning. However, while many visualization techniques for soccer have been developed in recent years, few have been directly applied to the video-based analysis of soccer matches. This paper provides a comprehensive overview and categorization of the methods developed for the video-based visual analysis of soccer matches. While identifying the advantages and disadvantages of the individual approaches, we identify and discuss open research questions, soon enabling analysts to develop winning strategies more efficiently, do rapid failure analysis or identify weaknesses in opposing teams. 
\end{abstract}


\copyrightyear{2019}
\acmYear{2019}
\acmConference[MMSports '19]{2nd International Workshop on Multimedia Content Analysis in Sports}{October 25, 2019}{Nice, France}
\acmBooktitle{2nd International Workshop on Multimedia Content Analysis in Sports (MMSports '19), October 25, 2019, Nice, France}
\acmPrice{15.00}
\acmDOI{10.1145/3347318.3355515}
\acmISBN{978-1-4503-6911-4/19/10}



%
% Keywords. The author(s) should pick words that accurately describe the work being
% presented. Separate the keywords with commas.

%
% A "teaser" image appears between the author and affiliation information and the body 
% of the document, and typically spans the page. 
%  \Description{Enjoying the baseball game from the third-base seats. Ichiro Suzuki preparing to bat.}

%
% This command processes the author and affiliation and title information and builds
% the first part of the formatted document.
\maketitle

\section{Introduction}

% 1. Introduce the general topic (informally).
The awareness of context is an important aspect for many applications incorporating and analyzing collective behavior data, of which team sports data is one rather specific but interesting example. One of the most famous team sports worldwide is soccer, with a large number of competitions held every year, many with huge financial and political investments and returns. Analyzing soccer matches to develop and implement winning strategies, do opposition research and failure analysis is relevant from an economic as well as a reputational perspective. Therefore, matches of national teams are often politicized and matter in a symbolic way, which further acts as a catalyst for larger financing (cf.~\cite{Machado.2017}). Working with collective behavior data is hard because the information and interrelations are rather complex. The data is, in general, a combination of hypervariate, hierarchical, relational, and temporal types. This leads to two challenges (cf.~\cite{Kraus.2019}). On the one side are the analytical aspects, which aim to find meaning inside the data, while on the other side is the visualization not only of the data itself, but also the information and knowledge that is derived from it. The latter enables us to support and enhance our understanding through meaningful representations and can give analytical insights, which help to reach the goals set out above.

A significant amount of research has been done on visualization techniques during the last decades. Meaningful progress has been made for the visualization of large multivariate, multidimensional, multitemporal, and spatial datasets. Information about and comparisons between them are available in the standard literature~\cite{Keim.2001, Keim.2002, Jin.2009, Oliveira.2003}. However, while some of the more generic approaches like tables, diagrams, and 2D maps have been applied to visualize sports data, few approaches have been developed directly with soccer in mind. Even fewer approaches exist that visualize relevant information \emph{in context}, i.e., directly embedded in a live stream or video recording, termed video-based visualizations. Video-based visualizations for soccer (and other team sports) have the advantage of augmenting information directly relevant to the actual scene. Furthermore, employed video-analysts by professional clubs are used to segment and analyze soccer matches manually based on available video recordings, which currently results in a very tedious and time-consuming process~\cite{Bialik.2014}. Visualization superimposed on the original video recordings ideally enable the addition of relevant information in real time during the match, directing the analysts attention to the essential aspects and events while maintaining the context to the real world, thus, enabling a user to form an extremely robust mental map (cf.~\cite{Basole.2016}). This, in turn, enables a more robust contextual understanding, makes forming connections between actions easier, and improves memorability.
  
% What is the problem?
Currently, most soccer match reports contain a significant number of statistics, presented mostly in large tables and only sometimes using (very simple) static graphics like shot distributions and set play analysis to visualize results. These visualizations typically include line diagrams, bar charts, positional plots, or star glyphs. Even when video-based visualizations are used in television recaps or summaries, they are usually created and placed manually or semi-automatic at best, which requires effort and carries significant cost with it. Furthermore, video-based approaches are mostly limited to track players, show offside positions or movements. Even these rather single-feature focused statistics and simple visualizations are an improvement of methods employed for the most parts of the last century, which involved simple counts (like number of successful passes, possession time, etc.) which have been read out or were later shown as overlay tables, with more detailed observations provided by subjective reports from experts. Recent progress in digital match tracking, computer graphics, and image manipulation has lead to increased capabilities in these regards, making advanced video-based visualizations technically feasible. However, the discipline is in its infancy, largely restricted to a commercial solution developed out of necessity, with only a few academic contributions so far, and much room for improvement still exists (cf.~\cite{Thomas.2017}).

% What do we propose (to close this gap, to foster resarch, ... )
In order to close this gap and improve as well as foster research in the field of video-based analysis of soccer matches, we provide a comprehensive and categorized overview of the latest, non-trivial methods developed for video-based visualization of soccer matches. We, specifically, do \emph{not} survey solutions which only provide tagging and enhanced video libraries like Hudl Sportscode~\cite{Sportscode.2019} or MyDartfish~\cite{Dartfish.2019}, as they do not employ video-based visualizations yet. However, plans in this regard are announced for the latter. In general, no current survey paper exists that describes the narrow field of video-based visualization of soccer matches completely. Existing surveys are either outdated~\cite{Borgo.2012} or incomplete~\cite{Perin.2018} as they describe the whole field of sports data visualization. Other surveys like~\cite{Rehman.2014, Shih.2018, Thomas.2017} can only be considered partial reviews (outdated, commercial solutions only, \textellipsis) or deal with a specific sub-issue. We survey the existing literature including outdated part-surveys (Section~\ref{sec:related_work}) and, furthermore, differentiate described video-based approaches from other visualization and analysis methods (Section~\ref{sec:methodology}). The existing approaches found in the literature are each presented shortly before we compare them according to eleven identified relevant features. We discuss differences and commonalities in detail, with a qualitative overview given in Table~\ref{tab:approaches_comparison} for reference (both Section~\ref{sec:survey}). Further we identify key issues and unsolved research challenges in the field and discuss research opportunities, before we provide a short summary of our findings (Section~\ref{sec:discussion}).

\section{Related Work}
\label{sec:related_work}
%1. Relate to current knowledge (What has been done?).
While a high volume of literature on visualization techniques exists~\cite{Keim.2001, Keim.2002, Jin.2009, Oliveira.2003}, which are extensively used in fields such as physics, the research is decidedly more sparse on the topic of video-based visualizations and in particular on video-based approaches for soccer.
The recent work by \citeauthor{Perin.2018}~\cite{Perin.2018} provides a quick overview of visualizations developed for sports data, although few techniques can be regarded as video-based, in particular for soccer. For those video-based approaches, an older survey by \citeauthor{Borgo.2012}~\cite{Borgo.2012} provides a general overview up to the year 2012, while developing useful classification schemes. However, with the fast pace of technological development, the referenced approaches are mostly outdated. Other surveys like \citet{Rehman.2014, Shih.2018, Thomas.2017} can only be considered partial reviews or deal with a specific sub-issue. They are either outdated, not complete, or consider only commercial approaches~\cite{Thomas.2007}.
Few solutions have been proposed in the field of video-based soccer visualization.
The first relevant work can be attributed to \citeauthor{Assfalg.2003}~\cite{Assfalg.2003} as early as 2003. It was one of the first to annotate soccer matches with calculated information, extracted from the video itself, to support the understanding of what is shown on screen. In the paper, ways are discussed to annotate videos with simple match scene information.
One of the most important contributors to the field, especially in the 2000s but also, later on, was Graham Thomas. His research as part of BBC~R\&D (British Broadcasting Company, Research \& Development) resulted in several commercial projects~\cite{iview.2005,augmentedplayer.2012, piero.2004}, which pioneered novel solutions for sports visualization, including the Piero system~\cite{piero.2018}, one of most widely used sports visualization and annotation system today and the main competitor to the Viz Libero system~\cite{vizlibero.2018}. Both systems are developed for sports broadcasting applications and provide a visualization and special effects toolbox, tailored to supported games like soccer. Both try to automate some aspects like tracking to reduce user workload.

In general, as detailed by \citeauthor{Thomas.2017}~\cite{Thomas.2017}, large commercial applications like Augmented Video Player~\cite{augmentedplayer.2012}, True View~\cite{trueview.2018}, Piero~\cite{piero.2018}, and Viz Libero~\cite{vizlibero.2018} for video-based soccer visualization are actually ahead of \emph{most} of the academic research and represent the state-of-the-art, with the only notable advanced academic exceptions being~\citet{Stein.2018, Stein.2018b}. Apart from these autonomously working solution, few other approaches with the same level of sophistication exist. The work by \citeauthor{Schlipsing.2014b}~\cite{Schlipsing.2014b} looked initially promising, but was abandoned after the principal investigator completed his Ph.D. Some other approaches exist, like~\citet{Wan.2004,Liu.2008, Xue.2017, Andrade.2005}. Nevertheless, these latter approaches remain at basic levels of visualization and surprisingly few new ideas were proposed. As described by \citeauthor{Thomas.2007}~\cite{Thomas.2007}, some individual and specific contributions have been made to the field of sports data analysis and visualization, however
\begin{quote}
	% do _not_ quote first part, as this is paraphrased for brevity and clarity
	\textit{whole systems which allow for automated tracking or labeling remain an open challenge (cf.~\cite{Thomas.2007}).}
\end{quote}
\section{Methodology}
\label{sec:methodology}

\textbf{Definitions and Terminology.}
Soccer, official known as \emph{Association football} and often--especially in Europe--just called \emph{football}, is a very popular team sport played between two competing teams of eleven players each. It is played on a grass field of rectangular size and the aim is to kick a spherical ball into the opponent's goal as often as possible, generally lasting 90 minutes in two parts. More detailed rules exist but are of no concern here. Games are very popular and several leagues, cups, and competitions have formed over time, often with huge financial impact. Games are often recorded as videos or transmitted live, depending on the venue, with several thousand or sometimes even millions of spectators. As one key to winning such games is understanding the opposing team and their tactic, past games are analyzed extensively. This process is currently done either manually or with semi-automated supportive technology and very time consuming as well as expensive \cite{Bialik.2014}. 
Here the automated match analysis comes in. As dedicated tracking devices can be expensive and more importantly would require the consent of the opponent teams, one is left with the only non-invasive technique remaining: video recordings of past and currently running matches.
This leaves video-based analysis as the only viable option for assessing an arbitrary soccer match.
The overall analysis process can be split into three different steps: extracting movement and event data from video data, analyzing this numeric and event data for assessments, and visualizing this information, either separately (e.g., as a statistics table) or reintegrate it into the video itself as a form of augmented reality. In this survey, we focus mostly on the latter part, but often this is impossible without considering the first two steps. Therefore, we also describe these aspects if necessary. The central aim of video-based soccer analysis can be regarded as searching for ways to visualize relevant match information inside the actual video. This can support a direct mental map, provide contextual information, and thereby enable coaches to significantly improve and simplify soccer match preparation and tactic planning. Video-based visualizations help to more effectively develop winning strategies, perform extensive failure analysis and identify weaknesses in opponents which are hard to detect otherwise. Therefore, we provide an overview of the existing literature which answers the following research question:
\begin{quote}
    \emph{Which techniques exist and are currently employed in video-based analysis of soccer matches?} 
\end{quote}

\noindent\textbf{Survey Structure.}
Papers are categorized into three groups: those that present a variety of approaches (mostly survey papers), those who present a single, new idea and those approaches which are described academically only rudimentary, as they are commercially available. For most cases, the approaches were either tested in reality (commercial solutions when developed and deployed) or, in case of academic research, have been evaluated as part of an expert study, to asses the practical significance of the new visualization. Interestingly, for most technical papers, the video-based visualization is only a by-product, with a larger part dedicated to analysis aspects. This is also apparent for the related work section as well as methodology, where mostly analytical approaches are referenced and analyzed. Therefore, less than a handful of papers describe in more detail the actual techniques employed and rationale behind the presented visualizations. Comparison between different video-based approaches, specifically for soccer, is virtually non-existent.
\section{Survey}
\label{sec:survey}
In the following Section~\ref{sec:survey_overview}, we give an overview of standard approaches and the most promising novel techniques proposed in literature and industry. In Section~\ref{sec:survey_comparison}, we propose classification methods and different criteria to assess the presented papers. We follow up with a detailed comparison of the presented approaches for each feature individually. A qualitative overview of the most important features of the approaches we consider most relevant is also given in Table~\ref{tab:approaches_comparison} for reference, which is presented after the approaches and comparison criteria.

%TODO Useful enumeration of techniques; Focus on the concepts and not on implementation details. [can be split into multiple chapters]

\subsection{Overview of current approaches}
\label{sec:survey_overview}
In the following, we present the individual approaches. The order in which the approaches are presented in is related to their \emph{complexity}. However, the details of this categorization can be safely ignored for now and the concept is introduced later, when relevant for the comparison in Section~\ref{sec:survey_comparison}.\newline

\begin{figure}
    \centering
    \includegraphics[width=\columnwidth]{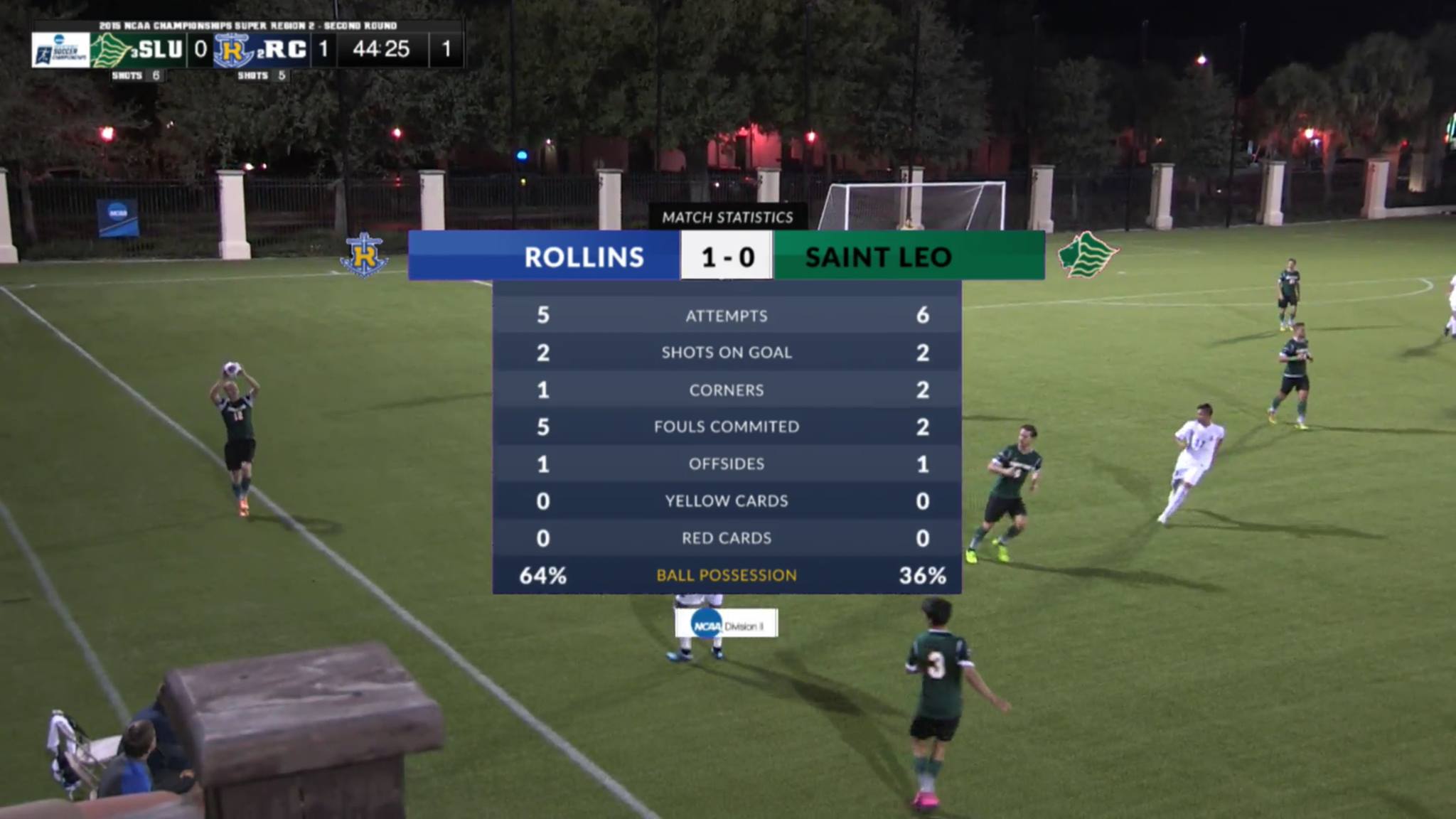}
    \caption{A table overlay showing current match statistics, while partly occluding the gameplay. The metrics display the relative strength of both teams and aim to explain the goal scored some seconds ago. Prepared by~\cite{SimpleThoughtProductions.2015}.}
    \label{fig:approach_table_overlay}
\end{figure}

\noindent\textbf{Basic: Textual/Numeric Table Overlay.}
The most simple form of video-based data visualization is at the same time one of the oldest. Here, the information, which is typicality text or numbers, is displayed as an overlay on top of the actual video, and shown for several seconds for it to be perceived. The only relation to the actual video is the timing, which can--but does not have to--match a current match situation. Information might include the current number of goals and some match statistics. An example of such a table overlay with some match statistics like goals, offsides, ball possession rate, etc. is shown in Figure~\ref{fig:approach_table_overlay}.\newline

\noindent\textbf{Automatic Highlight Identification.}
The approach by \citeauthor{Assfalg.2003}~\cite{Assfalg.2003} aims at automatically extracting highlights from a soccer game by analyzing the corresponding video sequences. First, playfield regions are extracted by their shape for localization. Then, player bounding boxes are found by color differentiating and adaptive template matching. To discern between different scenarios, the most likely scenario is chosen by using the normalized player position density distributions of example scenes. Then, a domain knowledge model computes the currently most likely scenario based on past predictions. The identified scenario is then reintegrated in the video in a bottom bar as event boxes for specific events, labeled as \emph{none}, \emph{hypothesis}, \emph{accepted}, and \emph{rejected}. While the detection technique was state-of-the-art at the time of publication and the accuracy was very good, the visualization was simple and did not evolve much. \newline
% effective study

\noindent\textbf{Highlights Extraction with Content Augmentation.}
The idea by \citeauthor{Wan.2004}~\cite{Wan.2004} differs somewhat from the previous aim by \citeauthor{Assfalg.2003}~\cite{Assfalg.2003}. Here, the first goal is to separate the video into different segments, using audio and visual cues, but only focusing on detecting replays. The second, more relevant goal is to identify regions for content augmentation. This is done by first determining the temporal and spatial relevance of the scene, as an insertion in a highly relevant scene would be disturbing. A suitable insertion area is then found by color quantization of homogeneous regions. No specific limits are set to the content to be inserted. However, as the position is in principle arbitrary and not related to any content, the most likely use is for non-obtrusive advertisement or other non-positional information. \newline
% effective study?

\noindent\textbf{Region-based Tracking and Provision of Augmented Information.}
The approach by \citeauthor{Andrade.2005}~\cite{Andrade.2005} differs from others by being primarily about extracting and tracking objects based on regions instead of single pixels. This region-based approach allows for a hierarchy of regions, represented as a graph structure. In turn, searching and tracking between different scenes are made more robust. Detected regions and features can the be augmented inside the video, by user-supplied graphics, although it should be noted that the primary focus of the work lies on the region-based tracking and the detection of semantic objects. \newline

\noindent\textbf{Virtual Content Insertion System.}
The work by \citeauthor{Liu.2008}~\cite{Liu.2008} is very similar to the second part of the one by \citeauthor{Wan.2004}~\cite{Wan.2004}. It proposes a generic Virtual Content Insertion system, which places content inside a video when the underlying actions attract much attention, but at a position where it does not obstruct the storyline. An adequate time is found by analyzing the spatial and temporal attention and the position using the three measures of saliency, contrast, and novelty. The insertion supports affine transformations to align the content with the actual environment. \newline
% effective study

\noindent\textbf{Automatic Video Annotation System.}
The system presented by \citeauthor{Xue.2017}~\cite{Xue.2017} aims at extracting metadata from video sequences of sports events. Sequence boundaries are segmented and clustered according to the type of content. Players cannot be tracked automatically but must be laboriously selected manually for tracking. On-screen text is recognized using trained SVM's (support vector machines) and read using OCR (optical character recognition). The combined information is presented as a simple on-screen graphic. \newline

%
% complex
%

\noindent\textbf{Video-based Activity Recording.}
In his thesis, \citeauthor{Schlipsing.2014b}~\cite{Schlipsing.2014b} presents a complete recording (calibrated dual-camera setup) and analysis system workflow for sports data analysis. Players are tracked semi-automatically using segmentation. Initial assignments and re-identifications need to be made manually. In terms of in-video visualization, player positions, and numbers, player formations (between manually selected players), and compactness hulls are presented. It was one of the first approaches which featured the placement of virtual objects in a soccer video for augmentation purposes. \newline

%\paragraph{Complex: StreamTeam}
%The approach StreamTeam by https://www.youtube.com/watch?v=QNSwX84voh4 \todo{}

\begin{figure}
    \centering
    \includegraphics[width=\columnwidth]{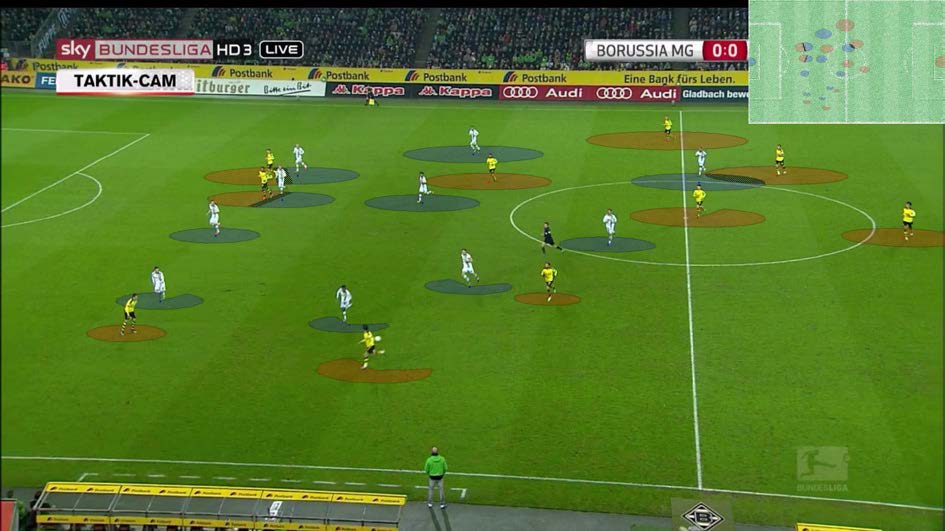}
    \caption{Interaction space visualization by \citeauthor{Stein.2018}~\cite{Stein.2018}. The augmented circles show the personal space that can easily be reached by the players and are helpful, for example, to identify suitable receivers for a pass.}
    \label{fig:approach_stein2018}
\end{figure}

\noindent\textbf{Combining Video and Movement Data to Enhance Team Sport Analysis.}
The approach presented by \citeauthor{Stein.2018}~\cite{Stein.2018} is somewhat similar to the main idea followed by \citeauthor{Schlipsing.2014b}~\cite{Schlipsing.2014b}, but differs in two fundamental ways in the implementation: First, it aims to use already existing video recordings--like a scouting feed or simple television recordings--as a video source, instead of a carefully calibrated dual-camera setup. Secondly, the tracking of the players aims to be autonomous and not require manual input. Then, in terms of visualization, the approach has a similar idea to enhance the gameplay and analysis by presenting visualizations in an augmented video. Here, several new ideas of features for visualization are presented: interaction spaces (the personal space that can be controlled by each player), free spaces, dominant regions, pass distances per player, and player reactions. An example visualization is shown in Figure~\ref{fig:approach_stein2018}. \newline

\noindent\textbf{Explanatory Storytelling for Advanced Team Sport Analysis.}
The follow-up paper to the previous approach, also by \citeauthor{Stein.2018b}~\cite{Stein.2018b}, proposes a system for automatic selection of helpful visualizations in key match situations. As part of this, two further visualization ideas are presented or improved upon: The pressure imposed upon players to act (aggressiveness of the defending team to recapture the ball) and an improved version of free/interaction spaces (spaces where nobody is standing or the personal space that can be reached by each player). \newline

%\paragraph{Commercial products}
\noindent The previously discussed approaches represent the state-of-the-art in the academic literature. In the following, we focus on commercial products that are deployed in the field.
According to \citet{Thomas.2017} and \citet{Hilton.2011}, the most widely used visualization products in the industry are \emph{Piero}~\cite{piero.2018}, developed by the BBC R\&D in collaboration with Ericsson, and \emph{True View}~\cite{trueview.2018} from Intel, former FreeD by Replay Technologies. Additionally, \emph{Viz Libero}~\cite{vizlibero.2018} is regarded as the third large broadcasting solution.
An overview of some of the systems is also given by \citet{Hilton.2011} and, more up to date, by \citet{Thomas.2017}. The commercial solutions are: \newline

\noindent\textbf{Commercial: iView (2005-2008).}
Developed by the BBC between 2005 and 2008, the iView: free-viewpoint video system~\cite{iview.2005} allows to reconstruct a 3D scene from the 2D video sequences and allows free positioning in them. In theory, arbitrary objects could be placed in the resulting 3D scene, but this was rarely done and the quality, as well as the resolution of the model, was very basic. The system was superseded by Intel True View (see below). \newline

\begin{figure*}[t!]
	\centering
	\subfigure[]{\includegraphics[width=0.49\linewidth]{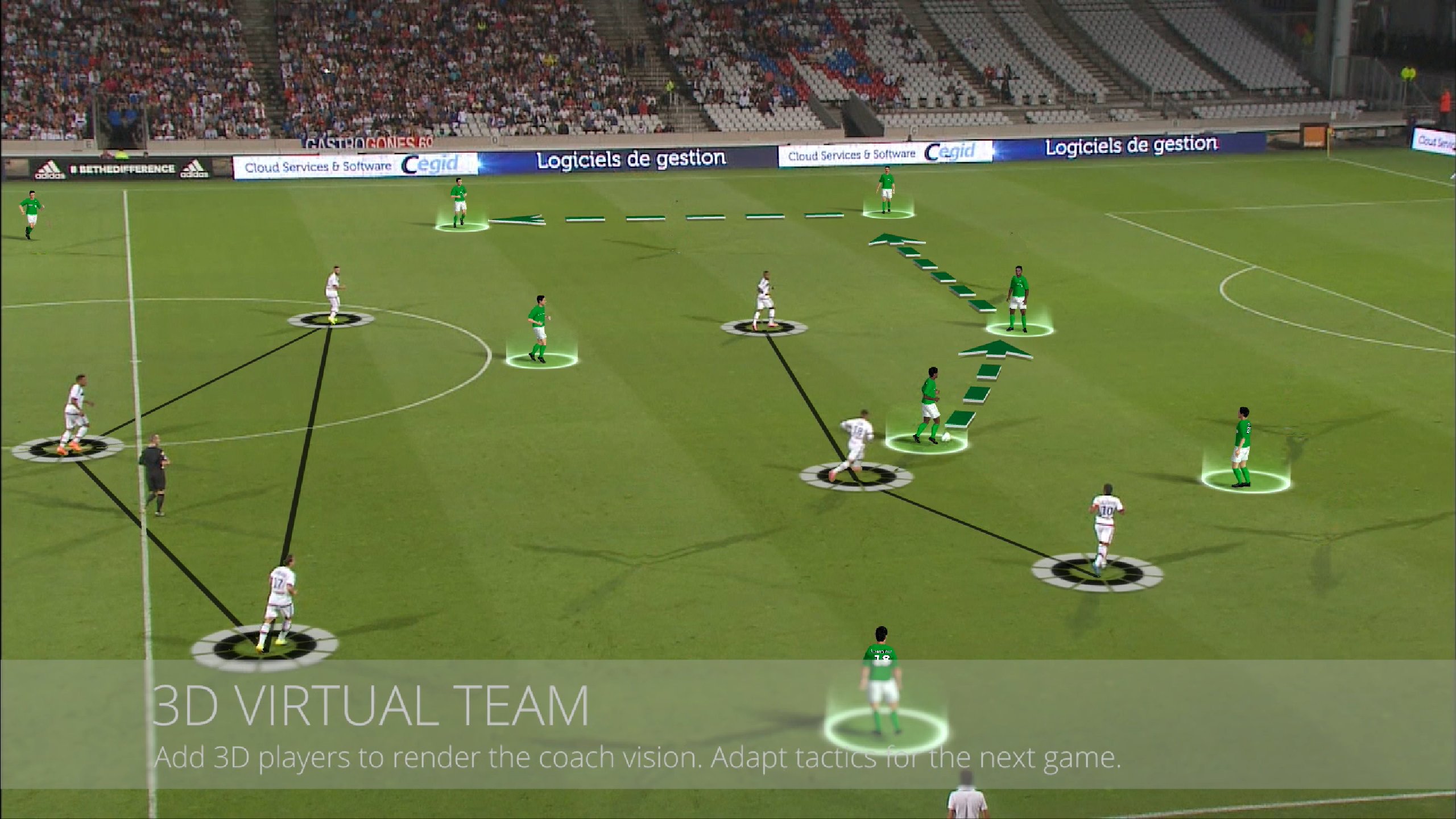}}\hspace{1mm}
	\subfigure[]{\includegraphics[width=0.49\linewidth]{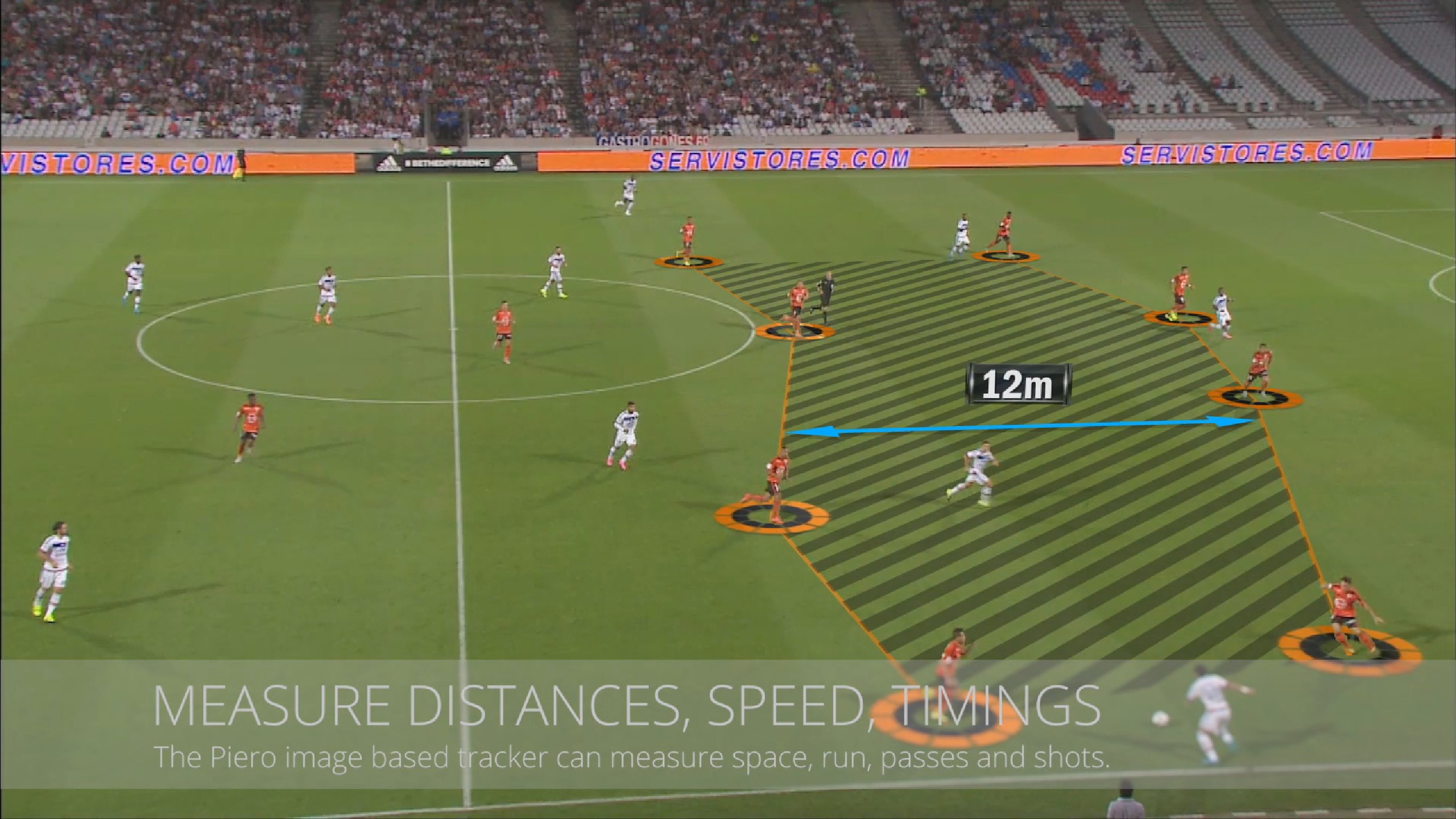}}\hspace{1mm}
	\subfigure[]{\includegraphics[width=0.49\linewidth]{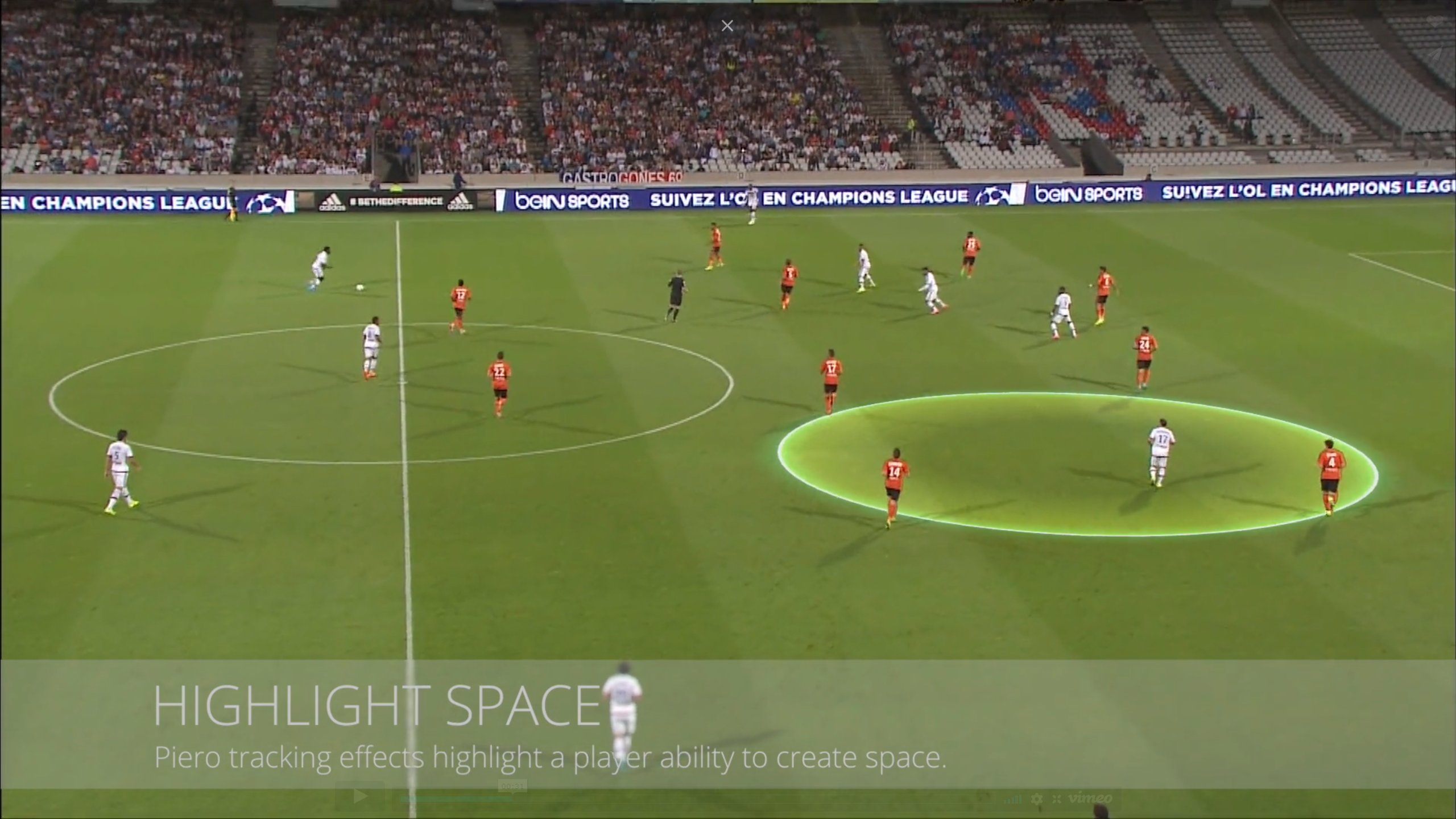}}\hspace{1mm}
	\subfigure[]{\includegraphics[width=0.49\linewidth]{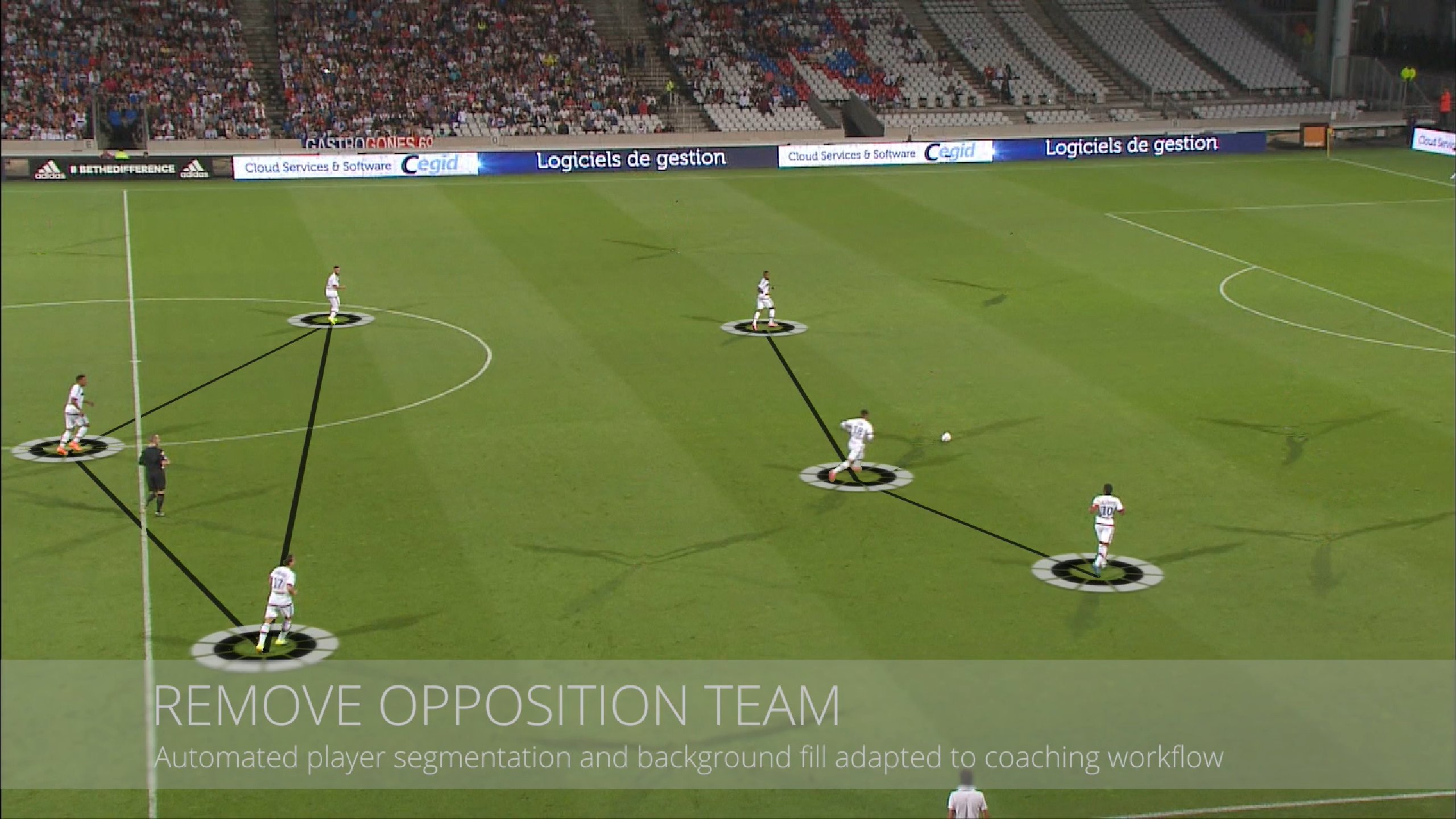}}
	\caption{Four different types of visualization which are available in the Piero System~\cite{piero.2018}.}
	\label{fig:piero_system_samples}
\end{figure*}

\noindent\textbf{Commercial: Augmented video Player (2012-today).}
Also developed by the BBC since 2012, the Augmented Video Player \cite{augmentedplayer.2012}, allows a user to watch sports sequences and interactively overlay additional layers to augment the video with further information. \newline

\noindent\textbf{Commercial: Intel True View (2012-today).}
The Intel True view system \cite{trueview.2018} is similar to the iView system and can be considered its technological successor. Formerly developed by FreeD Replay Technologies, it enables to reconstruct high-resolution 3D scenes using several HD cameras from different angles to show matches as a 3D scene. %(see also haimowitch 2015) \todo{}
The quality is markedly better and some simple (manually annotated) visualizations are used, like trajectories or highlighting. \newline

\noindent\textbf{Commercial: Piero (2004-today).}
The Piero system \cite{piero.2004, piero.2018} is considered one of the most advanced and widely used systems on the market. It was pioneered by the BBC, then outsourced to Redbeemedia in collaboration with Ericsson. It is used in different sports, including soccer, to augment the video with additional information. This aspect makes it come very close to the actual core of the survey. Tracking of players is done semi-automatically (initial assignment and re-identification are done manually) and several automatisms exist. However, the core of the systems remains a video-effects software, specifically tailored to real-time analytics of sports games. For soccer, several manually placeable visualizations are available: They include formations, free spaces, interaction spaces, distance measurements, ball trajectories, removable players, goal area heatmaps, player trajectories, density distributions, and others. Some of these examples are demonstrated in Figure~\ref{fig:piero_system_samples}. 
However, there is still a lot of manual work involved in creating animations and overlays. The main difference between the Piero system to the approach by \citeauthor{Stein.2018b}~\cite{Stein.2018b} is that the latter is a fully automated approach while the former represents the state-of-the-art in manually annotated video-based soccer analysis with an advanced graphical display. The visualizations by \citeauthor{Stein.2018b}~\cite{Stein.2018b} can also be considered advanced, but, more importantly, are based on mathematical and physiological considerations. However, sometimes they lack the clean and typographically nice style. Combining those two approaches would likely lead to enormous benefits. \newline

\begin{figure}
    \centering
    \includegraphics[width=\columnwidth]{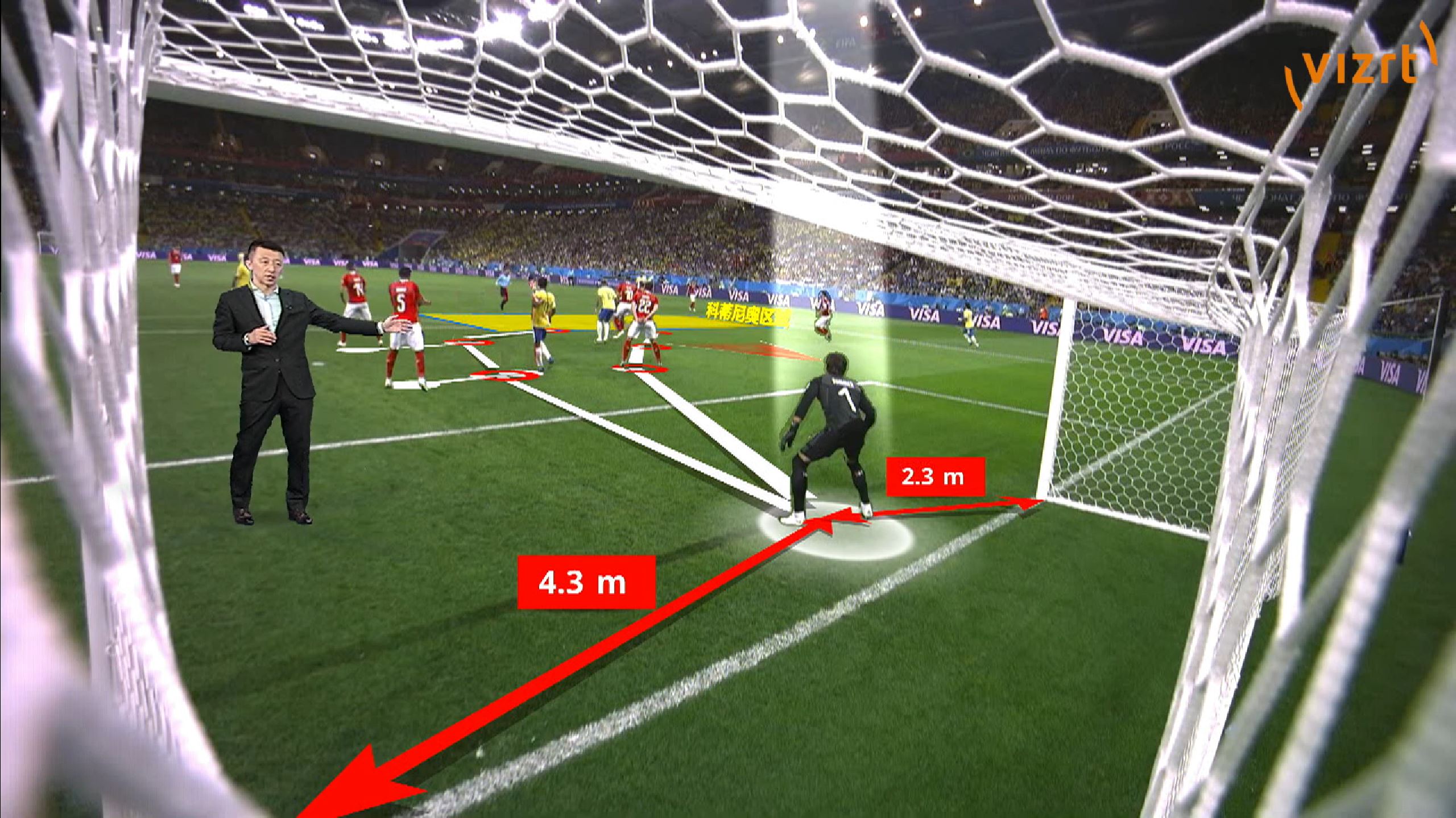}
    \caption{Annotated game scene with added distance measurements, highlighting and perspectively correct content placement by Viz Libero~\cite{vizlibero.2018}.}
    \label{fig:approach_viz_libero}
\end{figure}

\noindent\textbf{Commercial: Viz Libero (2010-today).}
The Viz Libero system~\cite{vizlibero.2018} is considered, together with the previously presented Piero system, the most advanced and widely used system on the market. It is also used for different sports, including soccer and augments the video with additional information. Available broadcast streams or video recordings are used as input streams, which means that no dedicated camera setup is required. Tracking is also semi-automatic and again several automatisms exist. The available visualizations are similar to the ones offered by Piero. One striking difference between Piero and Viz Libero is that the latter uses homography estimations and therefore allows limited reconstruction from single video sources. This means the camera viewpoint can be freely (within limits) be re-positioned within the scene to get the view from another perspective. One example of an augmented scene in Viz Libero is shown in Figure~\ref{fig:approach_viz_libero}.

\subsection{Categorization and Comparison}
\label{sec:survey_comparison} 
We first propose possible classifications and criteria to compare the different approaches. We establish the following nomenclature to classify the techniques into different categories. First, we differentiate between \textbf{\emph{scientific research}} and \textbf{\emph{commercial applications}}. For the latter, it is often the case that no paper exists that describes the system in detail, so one is left with product descriptions of the producer, manuals, as well as demo videos.
Secondly, we differentiate between the \textbf{\emph{complexity}} of the approaches, ranging from \textbf{\emph{low}} over \textbf{\emph{medium}} to \textbf{\emph{high}}. A special case is a simple data table overlay, which we assign to the class \textbf{\emph{very low}}, and is provided for reference as the simplest technique available.
To judge the amount of manual work required or how far an approach can work autonomous, we classify the proposals according to their \textbf{\emph{automation level}}, also ranging from \textbf{\emph{(very) low}} over \textbf{\emph{medium}} to \textbf{\emph{high}}.
In addition, we distinguish between the approaches according to several key concepts: Are they \textbf{\emph{insertion}} or \textbf{\emph{reconstruction}} techniques? Is augmented information contextually \textbf{\emph{embedded}} in the actual scene or merely overlayed and positioned? In this context, we further differentiate between static and support for \textbf{\emph{dynamic}} content as well as the support to \textbf{\emph{interact}} with the visualization, for example, by clicking or settings different parameters. Further options for comparison are if the content is \textbf{\emph{placed}} at fixed positions or positioned intelligently and if the system can in principle work in \textbf{\emph{realtime}} or not. To assess the impact of the research we further differentiate the papers and approaches by the presence or absence of a usability or case \textbf{\emph{study}}, either in the paper itself, as a follow-up paper, or in form of industrial applications. Lastly, it is relevant to note if the technique is still actively researched or even \textbf{\emph{used}} in the field.
A qualitative comparison of the important features from the presented approaches is given in Table~\ref{tab:approaches_comparison}. In this table, the individual features are grouped semantically, serving as an overview of the main findings of this survey. Sometimes the labels given in the table are over-simplified and do not give the whole aspect credit. The full evaluation is given in text-form in the following sections for each feature individually. \newline

\begin{table*}[ht!]
    \centering
    \ra{1.6}
    \begin{threeparttable}[t]
        \centering
        \setlength{\tabcolsep}{7pt}
        \arrayrulecolor{colorborder}
        \arrayrulewidth=.3pt
        \midsepremove
        \begin{tabular}{r|l|l|l|l|l|l|l|l|l|l|l|l|l|l|}
            %\arrayrulecolor{colorborder}
            % year
            \cmidrule[.3pt]{2-14}
            & \rotatebox[origin=l]{90}{\tc  1960+}
            & \rotatebox[origin=l]{90}{\tc  2003} 
            & \rotatebox[origin=l]{90}{\tc  2004}
            & \multicolumn{2}{c|}{\rotatebox[origin=l]{90}{\tc  2005} }   
            & \rotatebox[origin=l]{90}{\tc  2008}
            & \rotatebox[origin=l]{90}{\tc  2012}
            & \rotatebox[origin=l]{90}{\tc  2014}
            & \rotatebox[origin=l]{90}{\tc  2017}
            & \multicolumn{2}{c|}{\rotatebox[origin=l]{90}{\tc 2018} }   
            & \multicolumn{3}{c|}{\rotatebox[origin=l]{90}{\tc now} }    \\
            \cmidrule[.3pt]{2-14}
            
            % author / source
            & \rotatebox[origin=l]{90}{ \tsa Table Overlay}
            & \rotatebox[origin=l]{90}{ \tsa Assfalg     \cite{Assfalg.2003} }
            & \rotatebox[origin=l]{90}{ \tsa Wan         \cite{Wan.2004} }
            & \rotatebox[origin=l]{90}{ \tsa Andrade      \cite{Andrade.2005} }
            & \rotatebox[origin=l]{90}{ \tia iView\tnote{*} $\,$ \cite{iview.2005} }
            & \rotatebox[origin=l]{90}{ \tsa Liu         \cite{Liu.2008} }
            & \rotatebox[origin=l]{90}{ \tia AVP\tnote{*,3} $\;\;$ \cite{augmentedplayer.2012} }
            & \rotatebox[origin=l]{90}{ \tsa Schlipsing \cite{Schlipsing.2014b} }
            & \rotatebox[origin=l]{90}{ \tsa Xue          \cite{Xue.2017} }
            & \rotatebox[origin=l]{90}{ \tsa Stein\tnote{4} $\,$     \cite{Stein.2018} }
            & \rotatebox[origin=l]{90}{ \tsa Stein\tnote{4} $\,$     \cite{Stein.2018b} }
            & \rotatebox[origin=l]{90}{ \tia True View\tnote{*} $\,$     \cite{trueview.2018} }
            & \rotatebox[origin=l]{90}{ \tia Piero\tnote{*} $\,$        \cite{piero.2004, piero.2018} }
            & \rotatebox[origin=l]{90}{ \tia Viz Libero\tnote{*} $\,$        \cite{vizlibero.2018} } \midsepdefault \\
            % content
            \toprule
            %\midsepremove
            
            %\tc                                 &     &     &     &     &     &     &     &     &     &     &     &     &      \\ \cmidrule[.3pt]{1-14}\vspace*{0.3em}
            \tc \textbf{Complexity\tnote{1}}    &\tcz &\tca &\tca &\tca &\tcb &\tca &\tca &\tcb &\tca &\tcc &\tcc &\tcc &\tcc &\tcc  \\ \cmidrule[.3pt]{1-14}\addlinespace[1.2pt]\cmidrule[.3pt]{1-14}
            \tc \textbf{Insertion}                &\tml & \tm & \tm & \tm &     & \tm & \tm & \tm & \tm & \tm & \tm &     & \tm & \tm  \\ \cmidrule[.3pt]{1-14}
            \tc \textbf{Embedded}                &     &     &     &     &     & \tm &     & \tm &     & \tm & \tm &     & \tm & \tm  \\ \cmidrule[.3pt]{1-14}
            \tc \textbf{Reconstruction}         &     &     &     &     & \tm &     &     &     &     &     &     & \tm &     &  \tml \\ \cmidrule[.3pt]{1-14}\addlinespace[1.2pt]\cmidrule[.3pt]{1-14}
            \tc \textbf{Dynamic}                &     &     & \tm &     &     &     & \tm &     &     & \tm & \tm &     & \tm & \tm  \\ \cmidrule[.3pt]{1-14}
            \tc \textbf{Interactive}             &     &     &     &     & \tm &     & \tm &     &     &     &     & \tm & \tml & \tm \\ \cmidrule[.3pt]{1-14}
            \tc \textbf{Auto-Placement}         &     &     & \tm & \tm & \tm & \tm &     & \tm &     & \tm & \tm & \tm &\tml & \tml  \\ \cmidrule[.3pt]{1-14}
            \tc \textbf{Real-time}                 &     &     &     &     &\tml &     &     &     &     & \tm & \tm &\tml &\tml & \tml  \\ \cmidrule[.3pt]{1-14}\addlinespace[1.2pt]\cmidrule[.3pt]{1-14}
            \tc \textbf{User Study}             &     &     &\tml &     &     & \tm &     & \tm &     & \tm & \tm &\tml &\tml & \tml \\ \cmidrule[.3pt]{1-14}
            \tc \textbf{Active/Used}             & \tm &     &     &     &     &     &     &     &     & \tm & \tm & \tm & \tm & \tm \\ \cmidrule[.3pt]{1-14}\addlinespace[1.2pt]\cmidrule[.3pt]{1-14}
            \tc \textbf{Automation\tnote{1}}    &\tcz &\tca &\tca &\tca &\tca &\tca &\tca &\tcb &\tca &\tcc &\tcc &\tcb &\tcb & \tcb \\ 
            \bottomrule
        \end{tabular}
        \vspace{0.4em}
        \begin{tablenotes}
            \item[1] \crule[colorverylow]{6mm}{4mm} very low $\quad \quad$ \crule[colorlow]{6mm}{4mm} low $\quad \quad $ \crule[colormedium]{6mm}{4mm} medium $\quad$$\quad$ \crule[colorhigh]{6mm}{4mm} high
            \item[2] \crule[colornotmarked]{6mm}{4mm} not present $\quad$ \crule[colormarkedlight]{6mm}{4mm} partly $\quad$ \crule[colormarked]{6mm}{4mm} present  
            \item[3] Augmented Video Player
            \item[4] real-time capability in current, yet to be published versions
            \item[*] Commercial applications
        \end{tablenotes}
    \end{threeparttable}
    \caption{Comparison overview of the visualization techniques for video-based soccer match analysis we reviewed. Interesting to note is the low amount of current academic research and the relatively large amount of commercial solutions. For a detailed discussion, see Section~\ref{sec:survey_comparison}.}
    \label{tab:approaches_comparison}
\end{table*}

\noindent\textbf{Complexity.}
The most basic approach, with very low complexity, is just displaying a data table (or other types of simple visualization without location information) as an overlay on top of the video during the relevant period. The approaches with low complexity are all insertion techniques which employ non-sophisticated positioning inside the video and, with the exception of \citet{Liu.2008}, all operate in the screen space instead of the world space (embedding techniques). Several approaches of \citet{Assfalg.2003}, BBC R\&D (Augmented Video Player)~\cite{augmentedplayer.2012}, \citet{Andrade.2005}, \citet{Wan.2004}, \citet{Liu.2008}, \citet{Xue.2017} belong in this category. 
More sophisticated ideas with medium complexity are introduced by BBC R\&D (iView)~\cite{iview.2005} and \citet{Schlipsing.2014b}. The information is visualized as part of the scene itself.
The approaches Piero~\cite{piero.2004, piero.2018}, Viz Libero~\cite{vizlibero.2018}, True View~\cite{trueview.2018}, Stein et al.~\cite{Stein.2018,Stein.2018b}, with high complexity are those which employ advanced, embedded visualizations or reconstruction techniques, which require significant work to develop.\newline

\noindent\textbf{Automation.}
The automation level indicates how much user input is necessary for the analysis and to produce the visualization. Again, the simple approach of just displaying a data table contains \emph{very low automation}, as variables like placement, size, and duration all have to be determined manually. The vast amount of approaches like \citet{Assfalg.2003}, \citet{Wan.2004}, \citet{Andrade.2005}, iView~\cite{iview.2005}, \citet{Liu.2008}, Augmented Video Player~\cite{augmentedplayer.2012}, \citet{Xue.2017} offer \emph{low automation} and often several manual steps are necessary. The approach by \citet{Schlipsing.2014b} offers a higher degree of automatic visualizations, leading to \emph{medium automaton}. However, the initial selection of players for tracking and then the re-identification is still solved manually. The same is true for the systems True View~\cite{trueview.2018}, Piero~\cite{piero.2018}, and Viz Libero~\cite{vizlibero.2018}. These approaches all offer automatic support to reduce repeating and laborious steps. Nevertheless, the type of visualization and initial positions have to be specified manually. More often than not, even the actual content of the visualization is created manually by combining different building blocks. A \emph{high automation} level is only reached by the systems presented by Stein et al.~\cite{Stein.2018,Stein.2018b}. The visualizations are created automatically and without any user input, when using the default parameters.\newline

\noindent\textbf{Insertion.}
Techniques are considered insertion techniques when content information is placed inside the video, during a specific time and at a specific location. Most of the presented approaches belong to this category.
The exceptions are iView~\cite{iview.2005} and True View~\cite{trueview.2018}, which belong to the reconstructive techniques.\newline

\noindent\textbf{Embedded.}
This subcategory of insertion techniques contains approaches which actually embed the content into the 3D scene and respect relevant aspects like camera distortion and perspective projections. Approximately half of the approaches can be considered of this type, namely the ones by \citet{Liu.2008}, \citet{Schlipsing.2014b}, both from Stein et al.~\cite{Stein.2018,Stein.2018b}, Piero~\cite{piero.2018}, and Viz Libero~\cite{vizlibero.2018}. The four latter ones are the most sophisticated, leading to aesthetically pleasing results.\newline

\noindent\textbf{Reconstruction.}
Reconstructive techniques are not (primarily) based on inserting content into the video to help with the analysis, but reconstruct the scene as a 3D object to allow for arbitrary positioning. This can support match analysis by enabling different views. While the field of 3D reconstruction and structure from motion, part of image analysis and computer graphics, is relatively large, it has not primarily been applied to soccer, due to the unique challenges with tiny, moving targets. The exceptions are the commercial systems iView~\cite{iview.2005} and True View~\cite{trueview.2018}, which follow this reconstructive approach. True~View is the more advanced.\newline

\noindent\textbf{Dynamic Content.}
Depending on the type of content that can be inserted, the techniques can be classified to only support static, fixed size content like text and images, or they support dynamic content, which can be animated or morphed. From the presented approaches, only \citet{Wan.2004}, Stein et al.~\cite{Stein.2018,Stein.2018b}, \citet{augmentedplayer.2012}, Piero~\cite{piero.2018}, and \citet{vizlibero.2018} support this.

\noindent\textbf{Interactive.}
Most approaches output a generated augmented video file, which can not be interacted with. We consider a system to be interactive when, during playback, different parameters to modify the visualization or actually interact with it can be set by keyboard or mouse/touch. Of the presented techniques, only the reconstructive ones (iView~\cite{iview.2005}, True View~\cite{trueview.2018}), and Augmented Video Player~\cite{augmentedplayer.2012} are truly interactive. However, \citet{vizlibero.2018} can be considered to be mostly interactive and, to a lesser degree, this is also true for Piero~\cite{piero.2018}.
However, it should be said, that the boundaries between static and interactive are not strict and overlapping can exist. Also, some of the techniques can be adapted to provide interactivity.\newline

\noindent\textbf{Auto-Placement.}
The aspect of auto-placement is somewhat related to embedding, as it refers to the ability to place the content intelligently at an appropriate position. While this is a requirement for embedding, for overlaying techniques this is not strictly necessary, albeit often desired. Several of the presented systems are capable of automatically placing the content at the desired position, also different levels of accuracy and difficulty exist. While this is one of the main aspects for reconstructive techniques (merge different images for a 3D reconstruction), the approaches by \citet{Wan.2004}, \citet{Andrade.2005}, and \citet{Liu.2008} follow a simple positioning and only the approaches by \citet{Schlipsing.2014b} and Stein et al.~\cite{Stein.2018,Stein.2018b} are more advanced, with the latter doing a better job at the embedding.\newline

\noindent\textbf{Real-time.}
Of the presented approaches, only a few support real-time analysis and augmentation. For the purpose of this study, we consider any approach to be real time if the delay is below 50ms, a typical order up from when humans notice a lag. The reconstructive approaches can be considered near-real time with a delay in the order of seconds. This is achieved by not doing full 3D reconstruction but some forms of approximation yielding good enough results. For Piero~\cite{piero.2018}, the response time depends on the required human operations. The time needed to generate visualizations is often between several seconds to some minutes, with no upper limit in principle. \citet{vizlibero.2018} instead takes several minutes to generate non-trivial visualizations. Only the systems by Stein et al.~\cite{Stein.2018,Stein.2018b} support real time in their currently existing version.\newline

\noindent\textbf{Active/Used.}
Promising or successful systems are typically developed further or actively used. For the presented systems, to the best of our knowledge only the simple Table Overlay, the commercial systems True View~\cite{trueview.2018},  Piero~\cite{piero.2018}, and Viz Libero~\cite{vizlibero.2018} are used in the field. Further, the approaches by Stein et al.~\cite{Stein.2018,Stein.2018b} are actively developed and improved.\newline

\noindent\textbf{Study/Evaluation.}
Table Overlays have been in use for decades in the industry. These representations are extremely common and provide the basis for other, more detailed visualizations.
Studies discussing the other approaches only exists rudimentary for \citet{Wan.2004}, and as evaluations for \citet{Liu.2008}, Schlipsing et al.~\cite{Schlipsing.2014b}, as well as both paper by Stein et al.~\cite{Stein.2018,Stein.2018b}. For the industry applications True View~\cite{trueview.2018}, Piero~\cite{piero.2018}, and Viz Libero~\cite{vizlibero.2018} the usefulness is proven in form of continued usage and viewer feedback.
\newline

\noindent Concluding the comparison, we consider Piero~\cite{piero.2018} and Viz Libero~\cite{vizlibero.2018} the most advanced approach employed in the field in terms of visualization, while the research by Stein et al.~\cite{Stein.2018, Stein.2018b} represents the state-of-the-art in terms of automation of the whole process, generating the visualizations without user input and augmenting the video with relevant and useful information. 

\section{Discussion and Conclusion}
\label{sec:discussion}
Visualization aims to provide an abstract representation to enhance and speed-up the understanding of complex data. This is especially true for video-based soccer visualization, where the collective behavior of two groups of persons, trying to follow opposing tactics, leads to complex relational and hypervariate data, which is hard to visualize in context. An efficient and useful representation often is not known a priori, and more often than not depends largely on the specific aspects under consideration. This survey paper provided a comprehensive overview of the latest methods developed for the video-based analysis of soccer matches. The field is underdeveloped and few significant publications exist. The existing literature has been surveyed and relevant approaches have been identified. Apart from traditional ones like data tables, charts, and 2D maps, these include more advanced visualizations of formations, free spaces, interaction spaces, distance measurements, ball trajectories, selecting visible actors, player trajectories, and density distributions. The papers under consideration have been compared according to eleven criteria in the five areas \textbf{complexity}, \textbf{visualization type}, \textbf{capabilities}, \textbf{assessment}, and \textbf{automation level}. For all approaches, we have cited accompanying studies that show how each approach is expected to significantly improve and simplify soccer analysis, with some example scenarios given. We identify the approaches by Stein et al~\cite{Stein.2018,Stein.2018b}, the systems Piero~\cite{piero.2018} and Viz Libero~\cite{vizlibero.2018} as well as, in the area of 3D reconstruction for soccer, True View~\cite{trueview.2018} to be the most advanced methods currently available. In some sense, the main difference between the Piero system as well as the Viz Libero system compared to the approach by \citeauthor{Stein.2018b}~\cite{Stein.2018} is that the former more or less represent the state-of-the-art in manually annotated video-based soccer analysis with advanced graphical display, while the latter is a fully automated approach. Combining those two would likely lead to significant benefits. Almost all other academic approaches are in their infancy. This is surprising, given the technological advances in recent years and the commercial and political benefits which could be leveraged.

The process of analyzing soccer match performance and then representing the results in an understandable and meaningful manner, using video-based approaches, is a relevant topic which has not been adequately explored yet. The lack of approaches could be attributed either to a lack of research in this area, which might be considered more of an engineering than a scientific problem. However, this view falls short to respect that the academic literature has produced relevant publications of new types of visualization and the underlying mathematical definitions thereof, for example, for interaction spaces. Further, several open research questions exist, like aggregation and high-level representation of a game. It is a sign of significance that many of the presented systems are commercial solutions rather than research papers. This indicates that some works remain unpublished and are directly integrated into or developed for commercial products. A significant obstacle for scientific research is data unavailability, either due to legal or economic aspects. Nevertheless, given the ascent of big data analysis and machine learning in the last decade, and the increased awareness for performance analysis by the teams themselves, we expect research to intensify in the mid- and long-term, likely lead by private and company/club-internal research, which later will transpire to the academic community. Utilizing existing and proposed visualization techniques will provide great value to coaches in the areas of match analysis, game preparation, and opposition research. Simultaneously this will enable them to spot key issues otherwise gone unnoticed. With the emergence of increasingly detailed analysis of gameplay and tactics, visualization and, especially, in-video augmenting plays an increasingly important role. They help to understand capabilities and limits of a given team, identify and resolve specific issues and optimize the overall gameplay to adapt to new and complex challenges.

%
% The acknowledgments section is defined using the "acks" environment (and NOT an unnumbered section). This ensures
% the proper identification of the section in the article metadata, and the consistent spelling of the heading.
% TODO acknowledgments
%\begin{acks}
%	Here comes the acknowledgments. 
%\end{acks}

%
% The next two lines define the bibliography style to be used, and the bibliography file.
\bibliographystyle{ACM-Reference-Format}
\balance
\bibliography{bibliography}

% 
% If your work has an appendix, this is the place to put it.
%\appendix

\end{document}